\documentclass[letterpaper, 10 pt, conference]{ieeeconf}
\IEEEoverridecommandlockouts
\usepackage{booktabs}
\usepackage{xcolor}
\usepackage{epsfig}
\usepackage{graphicx}
\usepackage{amsmath}
\usepackage{amssymb}
\usepackage{subcaption}
\usepackage{pifont}

\newcommand{\cmark}{\ding{51}}%
\newcommand{\xmark}{\ding{55}}%

\usepackage{dsfont}
\usepackage{bbm}
\usepackage{algorithm}
\usepackage{algpseudocode}
\usepackage{bm}

\usepackage{xcolor,colortbl}
\usepackage{soul}
\definecolor{es-blue}{rgb}{0,0.4,0.8}
\definecolor{cvprblue}{rgb}{0.21,0.49,0.74}

\usepackage[breaklinks,colorlinks = true,citecolor=cvprblue, urlcolor=cvprblue]{hyperref}

\usepackage
[sortcites,
backend=bibtex,
bibstyle=ieee,
citestyle=numeric,
sorting = nyt,
mincitenames=1,
maxcitenames=2,
maxbibnames=15,
natbib=true,
doi=false,
isbn=false,
url=false,
eprint=false]{biblatex}

\makeatletter

\newcommand{\x}{\mathbf{x}}
\newcommand{\R}{\mathbb{R}}
\newcommand{\acronym}{GeFF}

\usepackage[margin=0.75in]{geometry}

\title{\LARGE \bf Learning Generalizable Feature Fields for Mobile Manipulation}
\author {
Ri-Zhao Qiu$^{*1}$,  Yafei Hu$^{*1,2}$, Yuchen Song$^{*1}$, Ge Yang$^{3}$, Yang Fu$^{1}$,  Jianglong Ye$^{1}$,  Jiteng Mu$^{1}$,  Ruihan Yang$^{1}$,
\\  Nikolay Atanasov$^{1}$,   Sebastian Scherer$^{2}$,   Xiaolong Wang$^{1}$ 
\\  $^{*}$equal contribution
\\  $^{1}$UC San Diego $^{2}$CMU $^{3}$MIT 
\\ {\color{es-blue}{\texttt{\url{https://geff-b1.github.io}}}}
}

\definecolor{es-blue}{rgb}{0,0.4,0.8}

\begin{document}

\twocolumn[{%
\renewcommand\twocolumn[1][]{#1}%
\maketitle

\vspace{-4mm}

\begin{center}
    \captionsetup{type=figure}
    \includegraphics[width=0.99\linewidth]{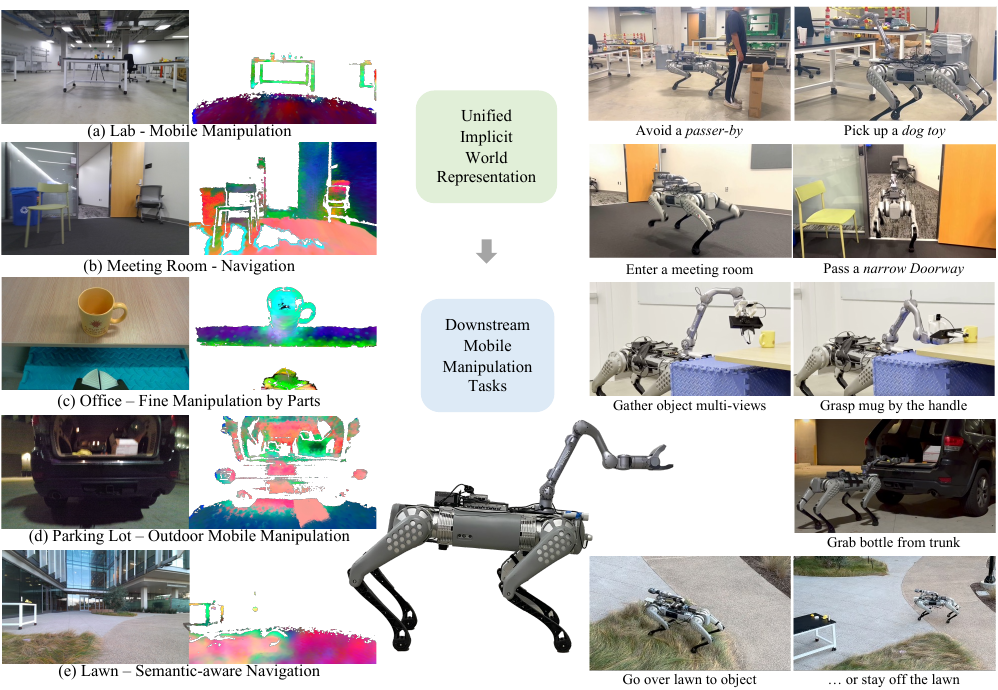}
    \caption{\small{\textbf{GeFF}, \textbf{Ge}neralizable \textbf{F}eature \textbf{F}ields, provide unified implicit scene representations for both robot navigation and manipulation in real-time. We demonstrate the efficacy of GeFF on {\bf open-world mobile manipulation}, {\bf semantic-aware navigation}, and {\bf zero-shot manipulation by parts} under diverse scenes ((a) work in a lab where a person walks in, (b) enter a meeting room with narrow entrance, (c) fine part-level manipulation, (d) grasp objects in a parking lot, and (e) semantic-aware navigation near a lawn).
    The visualization of the feature fields is obtained by PCA of rendered features.
    For best illustration, please check out the supplementary video.}}
    \label{fig:teaser_fig}
\end{center}%
}]

\begin{abstract}
An open problem in mobile manipulation is \textbf{\textit{how to represent objects and scenes}} in a \textbf{unified} manner so that robots can use both for navigation and manipulation. The latter requires capturing intricate geometry while understanding fine-grained semantics, whereas the former involves capturing the complexity inherent at an expansive physical scale.
In this work, we present \textbf{GeFF} (\underline{\textbf{Ge}}neralizable \underline{\textbf{F}}eature \underline{\textbf{F}}ields), a scene-level generalizable neural feature field that acts as a \textbf{\textit{unified}} representation for both navigation and manipulation that performs in \textbf{real-time}. To do so, we treat generative novel view synthesis as a pre-training task, and then align the resulting rich scene priors with natural language via CLIP feature distillation.
We demonstrate the effectiveness of this approach by deploying GeFF on a quadrupedal robot equipped with a manipulator. We quantitatively evaluate GeFF's ability for open-vocabulary object-/part-level manipulation and show that GeFF outperforms point-based baselines in runtime and storage-accuracy trade-offs, with qualitative examples of semantics-aware navigation and articulated object manipulation.
\end{abstract}

\section{Introduction}
	
Building a personal robot that can assist with common chores has been a long-standing goal of robotics~\citep{gupta2018robot, marques2023-TRINA, wu2023tidybot}. This paper studies the task of open-vocabulary mobile manipulation, where a robot needs to navigate through diverse scenes and manipulate objects based on language instructions. 
This task, while seemingly easy for humans, remains challenging for autonomous robots. Humans achieve such tasks by understanding the layout of rooms and the affordances of objects without explicitly memorizing every aspect. However, when it comes to robots, there does not exist a unified scene representation that captures geometry and semantics for navigation and manipulation tasks.

Recent approaches in navigation seek representations such as geometric maps (with semantic labels)~\cite{tian22tro_kimeramulti,Asgharivaskasi_ActiveMulticlassMapping_TRO23,qiu2022-RASLAM} and topological maps~\cite{shah2022gnm, shah2023vint} to handle large-scale scenes, but are not well integrated with manipulation requirements. Manipulation, on the other hand, often relies on dense scene representation such as implicit surfaces or meshes~\cite{wang2019stable,zhang2023-gamma,rashid2023-lerftogo} to compute precise grasping poses, which are not typically encoded in navigation representations. More importantly, supporting semantics-aware navigation with open-vocabulary object queries requires grounding to \textbf{geometric and semantic concepts} in the environment. The lack of a unified representation leads to unsatisfactory performance in open-vocabulary manipulation in large scenes~\cite{2023homerobot}. Performing coherent open-vocabulary perception for both navigation and manipulation remains a significant challenge.

We present a novel \textbf{\textit{scene-level}} \textbf{Ge}neralizable \textbf{F}eature \textbf{F}ield (\textbf{GeFF}) as a \textbf{\textit{unified}} representation for navigation and manipulation, trained with neural rendering akin to Neural Radiance Fields (NeRFs)~\cite{mildenhall2020nerf}. Instead of fitting a single static NeRF, GeFF only requires a single feed-forward pass to update the scene representation during inference. As a unified representation, GeFF stands out with two more advantages: (i) GeFF can decode multiple 3D scene representations from a posed RGB-D stream, including signed distance function (SDF) and pointcloud, and (ii) performing feature distillation from a pre-trained Vision-Language Model (VLM), \textit{e.g.}, CLIP~\cite{radford2021-CLIP}, GeFF provides language-conditioned semantics. Thus, GeFF mitigates the aforementioned discrepancy by supporting both real-time semantics-aware navigation ({\it e.g.,} avoiding humans) and zero-shot object part manipulation ({\it e.g.,} grasping mugs and tools by handles).

Using a quadrupedal mobile manipulator, we demonstrate that GeFF enables capabilities such as object-/part-level manipulation, semantics-aware navigation, and the potential to support articulated manipulation. We quantitatively show that GeFF outperforms existing point-based \cite{gu2023-conceptgraphs} and implicit \cite{Kerr2023-LERF} methods in open-vocabulary scene representation for mobile manipulation. Notably, the overall success rate \textbf{outperforms the best baseline by 19.2 absolute points on averaged object-level and part-level manipulation, while maintaining real-time efficiency}. In addition, we also qualitatively show that GeFF can be used to provide perception for other tasks such as semantics-aware navigation and articulated manipulation. We plan to release the pre-trained models and the source code.

\section{Related Work}
\textbf{Generalizable NeRFs.} Generalizable NeRFs extend conventional NeRFs' ability to render detailed novel views to scenes that come with just one or two images ~\cite{yu2021-pixelnerf, Trevithick2021GRF, wang2023-f2nerf,barron2023zip-nerf,wang2021-neus, varma2022-GNT, mu2023actorsnerf, ye2023-featurenerf}.
They replace the time-consuming per-scene optimization with a single feed-forward process through a network.
Existing work~\cite{varma2022-GNT, Tewari2021Advances, Rebain2022LOLNeRF} 
mainly focus on synthesizing novel views. Our focus is to use novel view synthesis via generalizable neural fields as a generative pre-training task. At test time, we use the produced network for representation generation on mobile robots.

\textbf{Feature Distillation in NeRF.}
Beyond just synthesizing novel views, recent work~\cite{Kerr2023-LERF,kobayashi2022-DFF,tschernezki2022-nerfdistill,ye2023-featurenerf} attempted to combine NeRF with feature distillation \cite{radford2021-CLIP,caron2021emerging,oquab2023-dinov2,rombach2022-latentdiffusion} to empower neural fields with semantic understanding of objects \cite{kobayashi2022-DFF,tschernezki2022-nerfdistill,ye2023-featurenerf}, scenes \cite{Kerr2023-LERF, shen2023-f3rm} and downstream robotic applications \cite{Ze2023GNFactor,shen2023-f3rm}.
PartSLIP~\cite{liu2023-partslip} and FeatureNerf~\cite{ye2023-featurenerf} performs part-level segmentation of objects, but require complete point clouds. Most closely related to our work, LERF-TOGO~\cite{rashid2023-lerftogo,Kerr2023-LERF} and F3RM~\cite{shen2023-f3rm} distill CLIP features for tabletop manipulation. We show that the conditional CLIP queries proposed in LERF-TOGO~\cite{rashid2023-lerftogo} apply to \acronym{} for part-based manipulation as well. Nonetheless, previous work cannot be easily adapted for mobile manipulation due to the expensive per-scene optimization scheme~\cite{Kerr2023-LERF,kobayashi2022-DFF} or restrictions to object-level representations \cite{ye2023-featurenerf}. In contrast, \acronym{} runs real-time on mobile robots.

\begin{figure*}[t]
    \centering
    \includegraphics[width=1.0\linewidth]
    {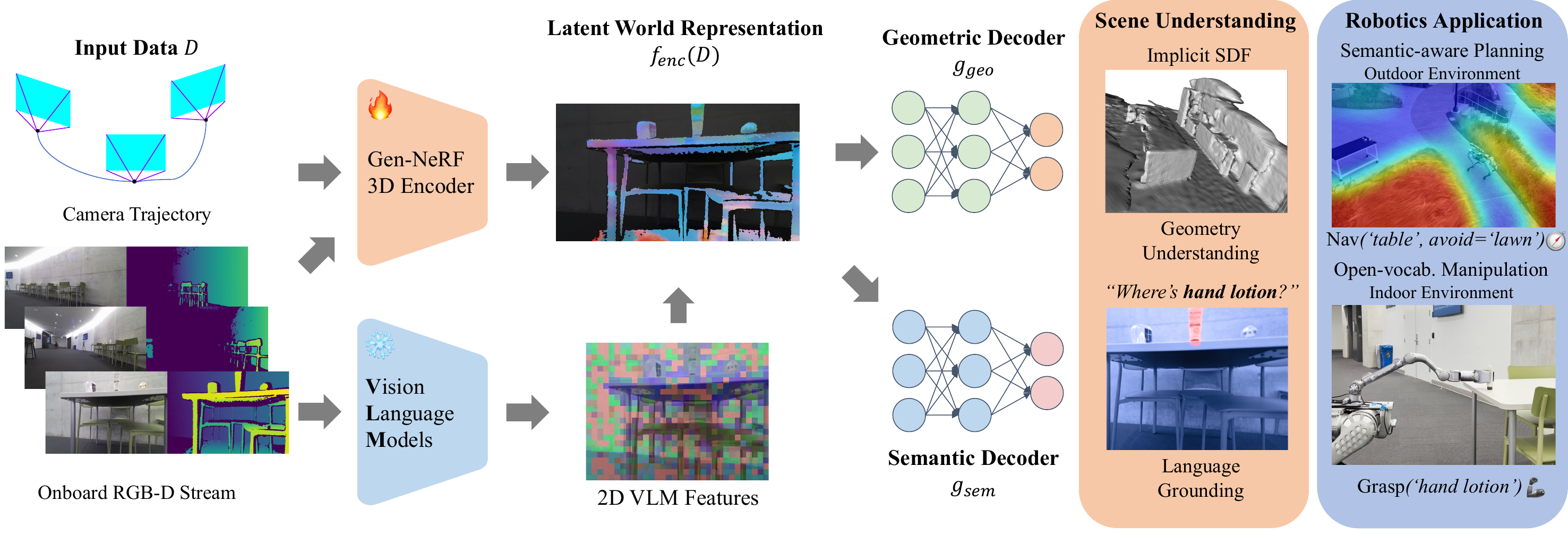}
    \caption{\small{Pre-trained as a generalizable NeRF encoder, \textbf{GeFF} provides a unified scene representation to support robot tasks from a onboard \mbox{RGB-D} stream, offering both real-time geometric information for planning and language-grounded semantics query capability. Compared to LERF~\cite{Kerr2023-LERF}, GeFF runs in real-time without costly per-scene optimization, which enables many potential robotics applications. We demonstrate the efficacy of GeFF in {\bf open-world language-conditioned mobile manipulation}. Feature visualizations are done by running PCA on high-dimensional feature vectors and normalizing the 3 main components as RGB.}}
    \label{fig:methodOverview}
    \vspace{-1em}
\end{figure*}

\textbf{Mobile Manipulation.} Besides work that perform closed-set mobile grasping~\cite{Kaelbling2012Unifying,Sun2021Fully,Wong2021Error,Gu2023Multi,Parashar2023SLAP,Xia2021ReLMoGen, 2023YokoyamaASC,Huang2023Skill,Stone2023moo,Blomqvist2020Go, Zimmermann2021Go,Parosi2023Kine}, there have been some recent work~\cite{liu2024-okrobot,chen2023-nlmap-saycan,huang2023-vlmap,jatavallabhula2023-conceptfusion,yokoyama2023-vlfm,gu2023-conceptgraphs,maggio2024-clio} that leverage 2D foundation vision models to for open-vocabulary mobile grasping and demonstration-based mobile manipulation~\cite{bharadhwaj2024-track2act}. Existing open-vocabulary manipulation methods project predictions from large-scale models~\cite{radford2021-CLIP,kirillov2023-segmentanything} directly onto explicit representations. This may require (1) offline optimization~\cite{gu2023-conceptgraphs}, expensive storage costs allowing only room-scale scenes and object-level grasping~\cite{gu2023-conceptgraphs,liu2024-okrobot}. GeFF, on the other hand, builds a \textit{latent and unified representation} for larger-scale outdoor environments and part-level grasping in real-time.

\section{GeFF for Mobile Manipulation}

\subsection{Problem Statement}

Given a coordinate $\x \in \R^3$ and a viewing direction $\mathbf{d}$ on the unit sphere $\mathbf{S}^2$, NeRF~\cite{mildenhall2020nerf} adopts an occupancy mapping $\sigma_{\theta}(\x): \R^3 \to [0, 1]$ and a color mapping $\mathbf{c}_{\omega}(\x, \mathbf{d}): \R^3 \times \mathbf{S}^2 \to \R^3$. Consider a ray $\mathbf{r}$ from a camera viewport with origin $\mathbf{o}$ and direction $\mathbf{d}$. NeRF estimates color along $\mathbf{r}$ by
\begin{equation}
    \mathbf{\hat{C}}(\mathbf{r}) = \int_{t_n}^{t_f} T(t) \alpha_{\theta}(\mathbf{r}(t)) \mathbf{c}_{\omega}(\mathbf{r}(t), \mathbf{d}) \mathrm{d}t\,,
\label{eq:color_rendering}
\end{equation}
where $t_n$ and $t_f$ are minimum and maximum bounding distances, $T(t) = \exp (-\int_{t_n}^t \sigma_{\theta}(s) \mathrm{d}s)$ is the transmittance capturing cumulative occupancy, and $\alpha_{\theta}(r(t))$ is the opacity value at $r(t)$ (in NeRF~\cite{mildenhall2020nerf}, $\alpha_{\theta} = \sigma_{\theta}$).

Let $\Omega$ be the space of RGB-D images.
Consider $N$ posed RGB-D frames $\mathcal{D} = \{(F_i, \mathbf{T}_i)\}_{i = 1}^{N}$, $F_i \in \Omega$, $\mathbf{T}_i \in \mathbf{SE}(3)$. Our goal is to create a \textit{\textbf{unified}} scene representation that captures geometric and semantic properties for robot loco-manipulation tasks. Specifically, we aim to design an encoding function $f_{enc}(\cdot): (\Omega \times \mathbf{SE}(3))^N \to \mathbb{R}^{N \times C}$ that compresses $\mathcal{D}$ to a latent representation, and decoding functions $g_{geo}(\cdot, \cdot): \mathbb{R}^3 \times \mathbb{R}^{N \times C} \to \mathbb{R}^m$ and $g_{sem}(\cdot, \cdot): \mathbb{R}^3 \times \mathbb{R}^{N \times C} \to \mathbb{R}^n$ that decode the latents into different geometric and semantic features at different positions in 3D space. The geometric and semantic features can then serve as input to a downstream planner. We aim to design these functions to meet the following criteria:

\begin{itemize}
\item {\bf Unified.} The encoded scene representation $f_{enc}(\mathcal{D})$ is \textit{\textbf{sufficient}} for both geometric and semantic query ({\it i.e.}, $g_{geo}$ and $g_{sem}$ are conditioned on $\mathcal{D}$ only via $f_{enc}(\mathcal{D})$).

\item {\bf Incremental.} The scene representation supports efficient incremental addition of new observations, ({\it i.e.}, $f_{enc}(\mathcal{D}_1 \cup \mathcal{D}_2) = f_{enc}(\mathcal{D}_1) \oplus f_{enc}(\mathcal{D}_2)$)

\item {\bf Implicit.} The encoded latents $f_{enc}(D)$ are organized in a sparse implicit representation to enable more efficient scaling to large scenes compared to storing $\mathcal{D}$.

\item {\bf Open-world.} The semantic knowledge from $g_{sem}$ is open-set and aligned with \textit{\textbf{language}}, so the robot can perform open-world perception.

\end{itemize}

We build GeFF upon generalizable NeRFs to satisfy these requirements. An overview of our method is shown in Fig.~\ref{fig:methodOverview}.

\subsection{Learning Scene Priors via Neural Synthesis}
\label{sec:gen_nerf_prior}

\begin{figure}[t]
    \vspace{-0.7em}
    \centering
    \includegraphics[width=\linewidth]{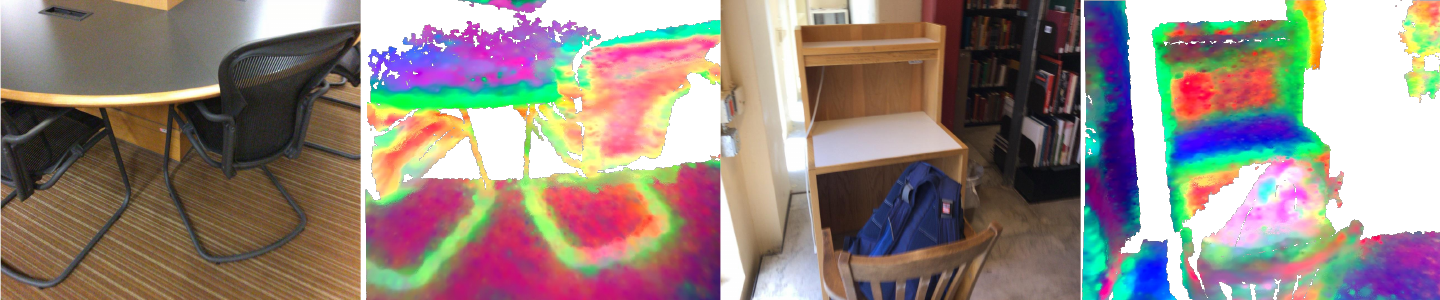}
    \caption{\small{\textbf{Generalizable NeRFs acquire geometric and semantic priors:} RGB images are input views from ScanNet~\cite{dai2017-scannet}, color images are PCA visualizations of feature volume projected to the input camera view encoded by an RGB-D Gen-NeRF~\cite{yangfu2023-sceneprior} encoder. Note how semantically similar structures acquire similar features.}}
    \label{fig:piloting_study}
    \vspace{-0.5em}
\end{figure}

Generalizable NeRFs (Gen-NeRFs) offer an effective pre-training objective for rich geometric and semantic priors~\cite{yangfu2023-sceneprior,huang2023-ponder,ye2023-featurenerf}. Fig.~\ref{fig:piloting_study} shows an illustration, rendering the latent feature volume from an RGB-D Gen-NeRF encoder~\cite{yangfu2023-sceneprior} trained to synthesize novel views on the ScanNet~\cite{dai2017-scannet} dataset. The colors correspond to the principal components of the latent features. We observe separations between objects and the background, despite that explicit semantic supervision was not provided during training.

GeFF uses two types of supervision to enhance these priors --- semantics using 2D features and geometry using SDF.

\textbf{Supervision (i): Language-Alignment via Feature Distillation.}\label{sec:language_alignment}
Although we have shown that Gen-NeRF encoders implicitly capture geometric and semantic cues, the representation is less useful if it is not \textit{\textbf{aligned}} to other feature modalities, such as language. To enhance the representation capability, in GeFF we use knowledge distillation to transfer learned priors from 2D vision foundation models and align the 3D representations with them. To the best of our knowledge, GeFF is the \textit{\textbf{first}} approach that combines scene-level generalizable NeRF with feature distillation. In contrast to previous works \cite{kobayashi2022-DFF,Kerr2023-LERF,ye2023-featurenerf}, which either require costly per-scene optimization \cite{kobayashi2022-DFF,Kerr2023-LERF} or is limited to object-centric representation \cite{ye2023-featurenerf}, GeFF both works in relatively large-scale environments and runs in real-time, making it a powerful perception method for mobile manipulation.

Specifically, we build a feature decoder $g_{sem}(\x, f_{enc}(D))$ on top of the latent representation, which maps a 3D coordinate to a feature vector. The output of $g_{sem}$ is trained to be aligned with the embedding space of a teacher 2D vision foundation model, termed $f_{teacher}$. Note that $g_{sem}$ is isotropic, as the semantics of an object should be view-independent regardless of the viewing directions. We can render 2D features for pre-training via
\begin{equation}
    \mathbf{\hat{F}}(r) = \int_{t_n}^{t_f} T(t) \alpha(r(t)) g_{sem}(\mathbf{r}(t), f_{enc}(\mathcal{D})) \mathrm{d}t\,,
\label{eq:feature_rendering}
\end{equation}%
which is modified from Eq.~\ref{eq:color_rendering}.
To further enhance the fidelity of the 3D scene representation, we use the 2D features of the input views computed by the teacher model as an auxiliary input to $f_{enc}$, which is
\begin{equation}
    f_{enc}(\mathcal{D}) = \textsc{ConCat}\left(\hat{f}_{enc}(\mathcal{D}), f_{teacher}(\mathcal{D}) \right)\,,
\label{eq:feature_rendering_encoder}
\end{equation}%
where $\hat{f}_{enc}$ is a trainable encoder and $f_{teacher}$ is a pre-trained vision model with frozen weights.
The final feature rendering loss is then given by standard L2 loss between $\hat{\mathbf{F}}$ and $\mathbf{F}$, which is the reference feature obtained by running $f_{teacher}$ on ground-truth novel views. Note that the input views and the rendered novel views are different adjacent views. The reference feature, $\mathbf{F}$, is obtained by running $f_{teacher}$ on ground-truth novel views.

\begin{figure}[t]
    \centering
    \includegraphics[width=\linewidth]{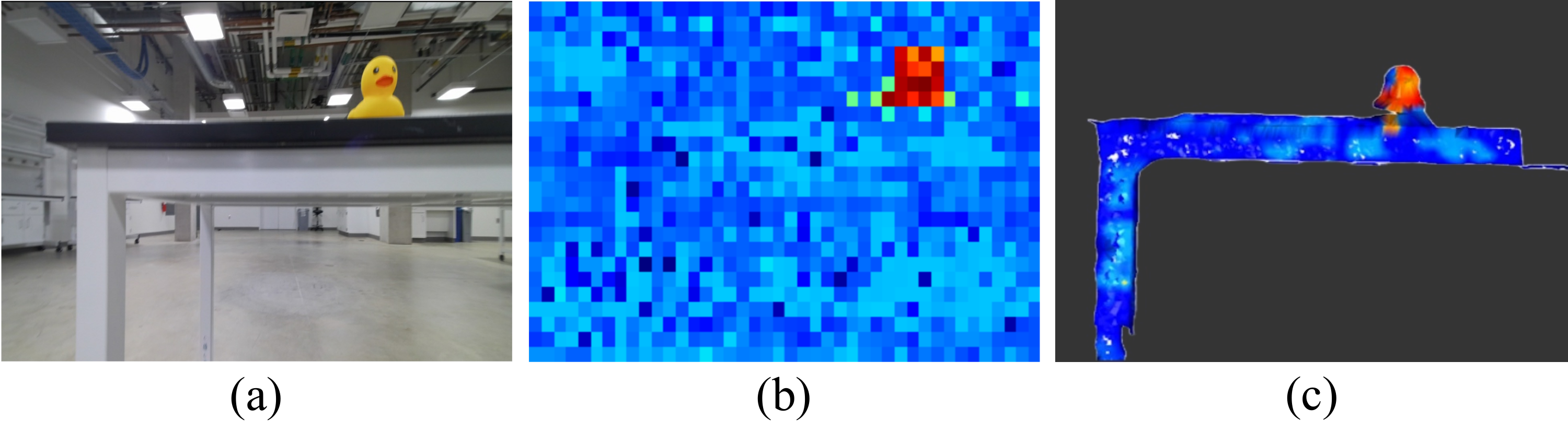}
    \caption{\small{\textbf{GeFF compresses and refines multi-view observations:} (a) single RGB view; (b) coarse 2D CLIP heatmap with query `toy duck'; (c) 3D heatmap from GeFF with clean boundary reconstructed from compressed latent representation.
    }}
    \label{fig:multi_view_feature_fusion}
    \vspace{-1em}
\end{figure}

\emph{Model for Distillation.} Our proposed feature distillation method for scene-level generalizable NeRFs is model-agnostic. In this work, since we are interested in \textit{open-vocabulary} tasks, we choose MaskCLIP~\cite{zhou2022-MASKCLIP} as $f_{teacher}$. MaskCLIP offers coarse (see Fig.~\ref{fig:multi_view_feature_fusion}) features but runs in real-time on mobile robots.

{\bf Supervision (ii): Depth Supervision via Neural SDF.} We use a signed distance network $s(\x) = g_{geo}(\x, f_{enc}(\mathcal{D}))$ to encode depth information, which is based on existing work~\cite{wang2021-neus,ortiz30-isdf,yangfu2023-sceneprior}. Doing so has two advantages over previous work~\cite{yu2021-pixelnerf}: 1) it leverages depth information to \textit{\textbf{efficiently}} resolve scale ambiguity for building scene-level representation, rather than restricted to object-level representation, and 2) it creates a continuous implicit SDF surface representation, which is a widely used representation for robotics applications such as computing collision cost in motion planning~\cite{ortiz30-isdf}.

To provide supervision for $g_{geo}$ during pre-training, we follow iSDF~\cite{ortiz30-isdf} and introduce an SDF loss $\mathcal{L}_{\text{sdf}}$ and an Eikonal regularization loss~\cite{gropp2020-eikonal} $\mathcal{L}_{\text{eik}}$ to ensure smooth SDF values. The main difference with iSDF~\cite{ortiz30-isdf} is that we condition $g_{geo}$ with $f_{enc}(\mathcal{D})$, which \textit{\textbf{does not require optimization for novel scenes}}. We represent the opacity function $\alpha$ in Eq.~\ref{eq:feature_rendering} using $s(\x)$%
\begin{equation}%
    \alpha(r(t)) = \textsc{Max} \left(\frac{\sigma_s(s(\x)) - \sigma_s(s(\x + \Delta))}{\sigma_s(s(\x))}, 0 \right)\,,
\end{equation}%
where $\sigma_s$ is a sigmoid with a learnable parameter $s$.
The depth along a ray $\mathbf{r}$ is then rendered by%
\begin{equation}%
    \mathbf{\hat{D}}(r) = \int_{t_n}^{t_f} T(t) \alpha(r(t)) d_i \mathrm{d}t\,,
\label{eq:depth_rendering}
\end{equation}%
where $d_i$ is the distance from the current ray marching position to the camera origin. Similar to Eq.~\ref{eq:feature_rendering}, the rendered depth can be supervised via standard L2 loss.

{\bf Final Training Objective.} Combining all the above equations, the total loss we used to train $f_{enc}$ for a unified latent scene representation is given by
\begin{equation}
    \mathcal{L} = \lambda_1\mathcal{L}_{col} + \lambda_2\mathcal{L}_{depth} + \lambda_3\mathcal{L}_{sdf} + \lambda_4\mathcal{L}_{eik} + \lambda_5\mathcal{L}_{feat}
\end{equation}
where $\lambda_i$s are hyperparameters used to balance loss scales. Empirically, we found that the feature quality is not sensitive to the choice of $\lambda$.
\definecolor{LightGray}{gray}{0.9}
\definecolor{asparagus}{rgb}{0.53, 0.66, 0.42}
\definecolor{applegreen}{rgb}{0.55, 0.71, 0.0}

\begin{table*}[t]
\centering
\caption{Open-vocabulary mobile manipulation success rate. Navigation success (Nav. Succ.) and composite mobile manipulation success (Mobile. Mani. Succ.) are reported for object-level tasks. For part-level tasks, we report manipulation success rates with different object-part queries ({\it e.g.,} mug-handle: grasping various mugs by handles). Latency represents the delay from the reception of the frame to the response of a text query on the onboard AGX Orin. The overall success is the average of overall success rates of object-level and part-level manipulation.
\textcolor{red}{$^{\star}$} methods require offline optimization with all observations batched together.} 
\label{table:main_results}
 \begin{tabular}{l c  c  c  c  c c c  c}
 \toprule
 & & \multicolumn{2}{c}{Object-level Mobile Manipulation} & \multicolumn{4}{c}{Part-level Manipulation} &
 \\
 \cmidrule(lr){3-4} \cmidrule(lr){5-8}
 Method &Latency & Nav. Succ. & Mobile Mani. Succ. & Mug-Handle & Tool-grip & Cart-bar & Avg. Succ. & Overall Succ.\\
 \midrule
 \acronym{} (Ours) & 0.39s
    & \textbf{94.4\%} & 61.1\% & {\bf 44.4\%} & {\bf 66.7\%} & {\bf 80.0\%} & {\bf 63.7\%} & {\bf 62.4\%}
 \\
 LERF\textcolor{red}{$^{\star}$}~\cite{Kerr2023-LERF} & $\sim$2 hrs\textcolor{red}{$^{\star}$}
    & 72.2\% & 44.4\% & 36.1\% & 20.0\% & 70.0\% & 42.0\% & 43.2\%
 \\
 ConceptGraph\textcolor{red}{$^{\star}$}~\cite{gu2023-conceptgraphs} & $\sim$200s\textcolor{red}{$^{\star}$}
    & \textbf{94.4\%} & \textbf{72.2\%} & 0\% & 20\% & 15\% & 11.6\% & 41.9\%
 \\
 ConceptGraph-Online & 4.63s
    & 5.56\% & 5.56\% & 0\% & 0\% & 15\% & 5\% & 5.3\%
 \\
 \bottomrule
 \end{tabular}
 \vspace{-1em} 
\end{table*}

\subsection{Implementing Open-Vocabulary Mobile Manipulation}
\label{sec:mobile_manipulation_applications}

{\bf Scene Mapping with \acronym{}.} \acronym{} encodes posed RGB-D frames to a latent 3D volume represented as a sparse latent point cloud, which can be built by concatenating per-frame observations. The camera poses are provided by an off-the-shelf VIO method~\cite{Seiskari2022HybVIO}.

{\bf Decoded Representations.} Though \acronym{} supports continuous decoding, it is inefficient to generate all possible representations densely on-the-fly. For this work, we decode the latent representation into discretized point clouds as geometric representations for navigation and manipulation. We then compute 2D grid by projecting the decoded 3D points and compute features for each grid cell by averaging the features of related points. This enhances basic units ({\it i.e.,} points and grid cells) with features from $g_{sem}$.

\textbf{Handling Language Query.} Following standard protocols~\cite{radford2021-CLIP}, \acronym{} takes in positive text queries and negative text queries ({\it e.g., ceiling}). To rate the language similarity, we compared decoded point features with text features using cosine similarity with a temperatured softmax. We sum up the probabilities of positive queries as the similarity score. For part-level language query, we use the conditional CLIP query technique proposed by \citet{rashid2023-lerftogo}. After the initial object is segmented, conditional CLIP query performs another pass of language query {\it conditioned on the segmented object with part-level prompt} for part segmentation.

{\bf GeFF for Navigation.} We consider the navigation of the quadruped robot as a 2D navigation problem following existing work~\cite{yokoyama2023-vlfm,2023homerobot,chaplot2020object}. Given text queries, we compare text embedding to grid embeddings. We use DBSCAN~\cite{ester1996-DBSCAN} to cluster high-response points for goal location and assign semantic affordances to grid cells. With an affordance-aware A$^*$ planner, this achieves semantic-aware navigation. Note that the 2D occupancy map is updated in real-time.

{\bf GeFF for Object-level Manipulation.} After the robot arrives at the goal receptacle, it searches for the target object by comparing semantics in points with given text, and uses DBSCAN to represent the target object as a centroid. In practice, we found that the parallel gripper has a high success rate in object-level grasping via an intuitive open-push-close gripper action sequence with trajectories computed by a sample-based planner (OMPL planner~\cite{sucan2012the-open-motion-planning-library}).

{\bf GeFF for Part-level Manipulation.} For objects that involve intricate geometry ({\it e.g.}, mug/tool with handles), it is counter-intuitive to solve the grasping problem with a centroid. In such cases, the user can provide specific parts to grasp via language. In \acronym{}, after the object centroid is localized, the robot can optionally use its in-wrist camera to gather multiple views, which adds millimeter-scale details to the representation. We then perform conditional CLIP queries and DBSCAN using significantly smaller EPS ({\it e.g.,} 1cm) to determine grasping location.

\section{Experiments}

\subsection{Experimental Setup}

\textbf{Training Details.} GeFF is pre-trained on the ScanNet dataset~\cite{dai2017-scannet}. for 50 epochs on a server with 8 RTX3090 GPUs in 6 days. We use the ViT-L CLIP model as $f_{teacher}$.

{\bf Robot Platforms.} We use the Unitree B1 as the base robot with a Unitree Z1 arm mounted on top of it. Besides a stereo camera and a structured light camera mounted at the robot head, the part-level experiments also uses an in-wrist camera to gather multi-view observations. The hardware setup can be seen in the supplementary video. 

\begin{table}[t]
  \centering
  \caption{\textbf{Ablation of auxiliary CLIP input} (Eq.~\ref{eq:feature_rendering_encoder}) on object-level mobile manipulation in diverse scenes. Navigation success rates (Navi.) and composite mobile manipulation success rates (Mani.) are reported.}
  \label{table:other_scenes}
  \resizebox{\linewidth}{!}{%
  \begin{tabular}{c c c c c c}
 \toprule
 & \multicolumn{2}{c}{Meeting Room} & \multicolumn{2}{c}{Kitchen} & 
 \\
 \cmidrule(lr){2-3} \cmidrule(lr){4-5}
 Methods & Navi. & Mani. & Navi. & Mani. & Overall
 \\
 \midrule
 \acronym{} (Ours)
    & \textbf{13/15} & \textbf{8/15} & \textbf{12/18} & \textbf{8/18} & \textbf{41/66}
 \\
 \acronym{} (no aux)
    & 9/15 & 5/15 & 7/18 & 4/18 & 25/66
 \\
 LERF~\cite{Kerr2023-LERF}
    & 6/15 & 3/15 & 8/18 & 5/18 & 22/66
 \\
 \bottomrule
\end{tabular}}
\vspace{-2em}
\end{table}

{\bf Real-world Evaluation.} For quantitative experiments, we use 4 environments: a 25$m^2$ lab , a 30$m^2$ meeting room, a 60$m^2$ community kitchen, and a 15$m^2$ office. For object-level experiments, unless otherwise noted, we use a total of 17 objects (6 misc., 5 office items, and 6 culinary items) including 8 novel categories that GeFF had \textit{\textbf{not seen}} during pre-training. For part-level manipulation, we use three different object categories with 4 instances each.

{\bf Experiment Protocol.} For all settings, we first manually drive the robot to build an initial representation of the scene to perceive receptacles (replaceable by standard robotic exploration algorithms). Then we provide task-related receptacle and object names to the robot.

\begin{figure*}[t]
    \centering
    \includegraphics[width=.99\linewidth]{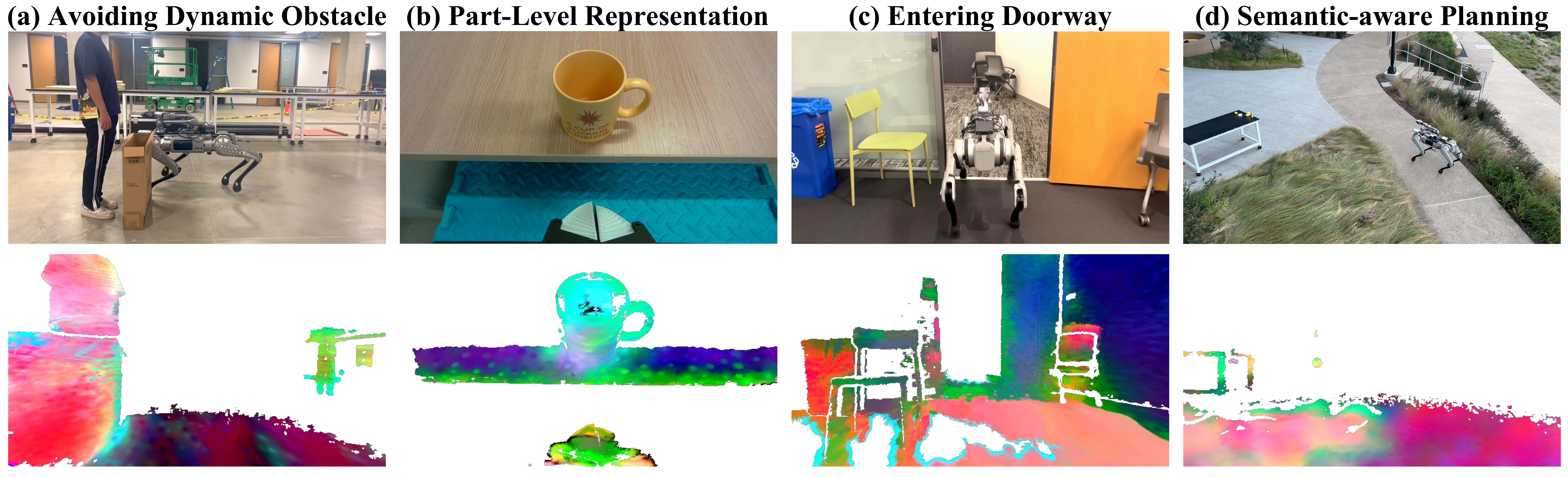}
    \caption{\textbf{Qualitative results of \acronym{} for diverse tasks:} (a) real-time update for dynamic person detection; (b) GeFF enables manipulation by parts; (c) entering a narrow doorway; (d) semantics-aware planning with affordance of \textit{`lawns'}. The results are animated in the supplementary video. Images in the second row are PCA visualization of first-person \acronym{} features.
    }
    \label{fig:mobile_manipu_qualitative}
    \vspace{-1em}
\end{figure*}

{\bf Baseline Implementation.} We choose two recent open-vocabulary scene representations as baselines. ConceptGraphs~\cite{gu2023-conceptgraphs} is a state-of-the-art open-vocabulary scene-level representation. Similar to OK-Robot~\cite{liu2024-okrobot}, it uses pre-trained vision models~\cite{liu2023-groundingdino,kirillov2023-segmentanything} for perception. Since both \textbf{ConceptGraph\textcolor{red}}{$^{\star}$} and \textbf{LERF\textcolor{red}{$^{\star}$}} require offline batch processing of all images, we process observed frames on a desktop computer. After which we manually provide object goals. \textbf{ConceptGraph-Online} is an online variant of CG, where it drops incoming frames if the previous frame is not finished processing. Since CG does not run on the AGX Orin, we re-use the same pipeline of ConceptGraph\textcolor{red}{$^{\star}$} but downsample the frames to match the latency. All representations are constructed by poses estimated by onboard VIO.

\begin{table}[t]
    \centering
    \caption{Mobile manipulation under {\bf scene change}, where objects are added after the initial scan. Note that methods~\cite{Kerr2023-LERF,gu2023-conceptgraphs} with expensive training requirement do not handle scene change.}
    \label{table:scene_change_results}
  \resizebox{\linewidth}{!}{
  \begin{tabular}{c  c  c   c   c}
 \toprule
 Method & Change & Lab & Meet. Rm. & Kitchen
 \\
 \midrule
 \acronym{} & \xmark
    & 7/9 & 7/9 & 8/9
 \\
 & \color{asparagus}{\cmark}
    & \cellcolor{asparagus}{4/9} & \cellcolor{asparagus}{6/9} & \cellcolor{asparagus}{8/9}
 \\
 \hline
 LERF~\cite{Kerr2023-LERF} & \xmark
    & 6/9 & 7/9 & 4/9
 \\
 & \color{asparagus}{\cmark}
    & \cellcolor{asparagus}{NA$^*$} & \cellcolor{asparagus}{NA$^*$} & \cellcolor{asparagus}{NA$^*$}
 \\
 \bottomrule
\end{tabular}}
\vspace{-2em}
\end{table}

\subsection{Evaluation}

We answer important \textbf{R}esearch \textbf{Q}uestions: How is \acronym{} compared to other open-vocabulary scene representation methods (A1, A2, A3, A4)? How is \acronym{} compared to simple projection baseline (A6)? What were the design choices (A5)? Can GeFF be used for diverse tasks (A7)?

\noindent{\textbf{A1. ConceptGraph requires offline optimization and breaks when real-time requirement is enforced.}} From Tab.~\ref{table:main_results}, we can see that ConceptGraph works at the cost of \underline{expensive offline processing}, which is not suitable for mobile robots. When ConceptGraph is granted offline processing using desktop-level compute, it achieves \textit{slightly} better results than GeFF on object-level grasping. However, when it is forced to perform online inference, we empirically observe its internal point cloud merging design breaks due to its assumption of adjacent frame proximity, which leads to degenerate representations and bad success rate.

\noindent{\textbf{A2. ConceptGraph fails to respond to part-level queries.}} Specifically designed for object-level representations, ConceptGraph can not support part-level grasping ({\it e.g.,} grasping a screwdriver by handle instead of shank), which is evident from Tab.~\ref{table:main_results}. Specifically, it generates no or bad responses to part-level queries such as {\it handles} or {\it grips}, which is due to lack of part-level training data in the open-vocabulary detector~\cite{liu2023-groundingdino} that ConceptGraph relies on.

\begin{table}[t]
    \centering
    \caption{GeFF learns geometric priors to reconstruct geometry from compressed latent representation. Both GeFF and projection baselines downsample the depth and MaskCLIP features to at most 512 points. Depths are reconstrcuted/upsampled and compared to reference depth.}
    \label{tab:geom_quantitative}  
    \resizebox{\linewidth}{!}{%
    \begin{tabular}{l|c}
    \hline
    Method                              & Depth L2 Error$\downarrow$ \\ \hline
    GeFF                                & \textbf{0.012}          \\
    Projection (Nearest interpolation)  & 0.061          \\
    Projection (Bilinear interpolation) & 0.040         \\ \hline
    \end{tabular}}
\vspace{-2em}
\end{table}

\noindent{\textbf{A3. Unlike GeFF, LERF requires offline processing and does not provide clear boundary.}} LERF~\cite{Kerr2023-LERF}, another feature field method, is an RGB-only method with view-dependent features. Thus we select the point with maximum responses in features rendered from training views as the goal location. Due to lack of geometric supervision, \textit{LERF often fails due to (1) noisy responses from under-observed areas and (2) unclear object boundaries}. However, as a continuous implicit method, \textbf{LERF show significantly better performance on part-level manipulation than ConceptGraph}, which is consistent with our finding that continuous representation is better suited for part-level representation.

\noindent{\textbf{A4. GeFF works when scene changes with slightly worse performance.}} For manipulation under \textbf{scene change}, we place a subset of objects (hand lotion, bottle, dog toy) on the table {\it after} the initial scan with 3 trials each. Tab.~\ref{table:scene_change_results} shows the results. Both LERF~\cite{Kerr2023-LERF} and CG~\cite{gu2023-conceptgraphs} are not applicable for scene changes as they require costly re-training. One potential cause for the decrease is the lack of multi-view observations as the robot only gets a front view when it approaches the receptacle.

\noindent{\textbf{A5. Auxiliary 2D input helps with generalization.}} We ablate \acronym{} the effectiveness of Eq.~\ref{eq:feature_rendering_encoder} in more diverse environments in Tab.~\ref{table:other_scenes}. Specifically, we found that, if auxiliary input is not used, \acronym{} shows decreased performance especially on objects absent from pre-training on ScanNet~\cite{dai2017-scannet}. We believe that auxiliary input provides a `shortcut' generalization beyond training data, which may replaced by a significantly larger training scale.

\noindent{\textbf{A6. The learned geometric priors are effective at compression.}} To evaluate the learned geometric priors, we reconstruct depth from the latent representation to compare it with reference depth. For a given RGBD frame, GeFF encodes it to 512 latent points and reconstructs the depth. The simple projection baseline downsamples the given RGBD frame to 512 pixels, and interpolates back to the original resolution. The resulting L2 errors between reconstructed depths and reference depths are given in Tab.~\ref{tab:geom_quantitative} using 10 validation scenes of the ScanNet dataset, where GeFF shows significantly better geometric error.

\noindent{\textbf{A7. GeFF can serve as the 3D perception backbone for diverse tasks.}} We show qualitatively in both Fig.~\ref{fig:mobile_manipu_qualitative} and the supplementary material that GeFF features are fine-grained and real-time enough to perform diverse tasks beyond grasping, such as \textbf{dynamic obstacle avoidance, semantic-aware navigation, and articulated manipulation for door opening}, which highlights its potential to provide 3D representation for robotics tasks.

\section{Conclusion}
In this paper, we present GeFF, a scene-level generalizable neural feature field with feature distillation from VLM that provides a unified representation for robot navigation and manipulation. Deployed on a quadruped robot with a manipulator, GeFF demonstrates zero-shot object retrieval ability in real-time in real-world environments. Using common motion planners and controllers powered by GeFF, we show competitive results in open-set mobile manipulation tasks.

\clearpage
\printbibliography

\end{document}